# Thread Detection and Response Generation using Transformers with Prompt Optimisation

Kevin Joshua T
School of Electronics Engineering
Vellore Institute of Technology, Chennai
600127, India
kevinjoshua.t2021@vitstudent.ac.in

Arnav Agarwal
School of Electronics Engineering
Vellore Institute of Technology, Chennai
600127, India
arnav.agarwal2021c@vitstudent.ac.in

Shriya Sanjay
School of Computer Science and Engineering
Vellore Institute of Technology, Chennai
600127, India
shriyasanjay.nari2021@vitstudent.ac.in

Yash Sarda
School of Electronics Engineering
Vellore Institute of Technology, Chennai
600127, India
yashpavankumar.sarda2021@vitstudent.ac.in

John Sahaya Rani Alex
School of Electronics Engineering
Vellore Institute of Technology, Chennai
600127, India
jsranialex@vit.ac.in

Saurav Gupta
School of Electronics Engineering
Vellore Institute of Technology, Chennai
600127, India
saurav.gupta@vit.ac.in

Sushant Kumar
Samsung R&D Institute India, Bangalore
560037, India
sushant.kr@samsung.com

Vishwanath Kamath
Samsung R&D Institute India, Bangalore
560037, India
vishwanath.p@samsung.com

*Abstract*—Conversational systems are crucial for human-computer interaction, managing complex dialogues by identifying threads and prioritising responses. This is especially vital in multi-party conversations, where precise identification of threads and strategic response prioritisation ensure efficient dialogue management. To address these challenges an end-to-end model that identifies threads and prioritises their response generation based on the importance was developed, involving a systematic decomposition of the problem into discrete components - thread detection, prioritisation, and performance optimisation which was meticulously analysed and optimised. These refined components seamlessly integrate into a unified framework, in conversational systems. Llama2 7b is used due to its high level of generalisation but the system can be updated with any open source Large Language Model(LLM). The computational capabilities of the Llama2 model was augmented by using fine tuning methods and strategic prompting techniques to optimise the model's performance, reducing computational time and increasing the accuracy of the model. The model achieves up to 10x speed improvement, while generating more coherent results compared to existing models.

## I. INTRODUCTION

The art of seamlessly integrating computers into our daily lives without any sort of barrier has attracted increasing attention due to its forecasted future and promising commercial viability. In the initial phase, many researchers focused on building dialogue generation models [1][2] with various neural networks primarily on verbal interactions between two individuals. This changed over time to focus on a realistic and complex scenario of conversations with multiple participants - multi-party conversations (MPCs) [3] .

Utterances in a two-party conversation are posted one by one between two interlocutors, constituting a *sequential* information flow. Utterances in an MPC can be spoken by anyone and address anyone else in the conversation. The goal of conversation disentanglement is to break down mixed messages into separate conversations. However, current disentangling methods are largely based on manual, dataset specific features, restricting generalisation and flexibility.

In this paper, a comprehensive framework for conversation disentanglement that eliminates the need for domain-specific, time-consuming feature engineering, an optimal approach to embed the entire utterance (timestamp, speaker, and text) is proposed which is more context specific and conversation oriented, predicting ideal responses better than other models.

The model forms threads using inherent linguistic understanding of the Transformer based LLMs, while reducing the computational time by abstracting away the process of output generation while classifying threads. The efficiency has been enhanced for optimal performance on consumer hardware, providing real-time results.

## II. RELATED WORKS AND BASELINE MODELS

While evaluating current State Of The Art models, it was found that researchers arrived at the same conclusion, even with different analytical methods. Identification of speaker, addressee, context of the content, and time, influenced the quality of response. A stacked Long Short Term Memory(LSTM) model, Bidirectional Encoder Representations from Transformers (BERT)[5][6][7] models and a pointer network based approach are used as baseline models.

The Stacked LSTM model [4], characterised by multiple LSTM layers that are arranged one on top of another, necessitates three-dimensional input, while LSTMs generate a two-dimensional output, reflecting an interpretation derived from the end of the sequence. This shows the dependence of the model on time as an evaluation metric for the development of the response by the model.

The BERT based model [8], addresses challenges in dialogue state tracking (DST) by focusing on scalability for dynamic ontologies and unseen slot values. Unlike prior approaches that rely on candidate generation or slot tagging, it is an end-to-end DST model that directly extracts slot values from the dialogue context using BERT, a powerful contextual language model. It leverages BERT's contextualised representations to predict slot values as "none," "don't care," or text spans within the dialogue context. Additionally, the model employs parameter sharing across all slots to reduce the number of parameters and facilitate knowledge transfer.

In the pointer based model [9] as given in Fig. 1., disentanglement of chats is done in a non domain-specific fashion, which has been an increasingly challenging task among other generic models. This is done by considering it as a pointing problem, where the model learns to point from parent utterance and is modelled as a multinomial distribution over a set of previous utterances.

Each of the utterances is characterised by the text, time stamp and speaker to link between the conversations which helps in removing the main issues hindering the accuracy in generic models- self linking. Self linking is an issue that arises in two conditions- start of conversation and isolated messages. In both the scenarios, there is no proper reply or start to the message to be linked with, resulting in the breaking of positively linked clusters and hindering accuracy. This is the very first work that utilises a pointer network for conversation disentanglement and shows better accuracy in cluster predictions and forming link predictions compared to several other models due to its efficient handling of absence of messages and proper topic coherence formation.

The transformer and heterogeneous graph network model [10] performed with a high level of accuracy in the *UBUNTU* dataset. However due to long input lengths and inefficiencies of prompting, the response takes a long period of time to generate. It also calculates addresser and addressee which may not always be necessary in this context and thread id is more relevant in most cases. Additionally the use of interlocutors establishes 2 kinds of nodes in the graph that are utilised by the network as well.

The model incorporating Adam Optimiser and TSL [11] aims to address the challenge of disentangling online dialogues. The model captures sequential information during the online algorithm's processing of a dialogue, enabling effective disentanglement of dialogues. The coherence within each session is maintained as the algorithm proceeds, enhancing the model's ability to discern and separate intertwined conversations online. It has the ability to understand the chronological order of utterances and improves overall disentanglement performance. It is a strong baseline due to its innovative architecture, practical implementation, and recognition in academic literature with a high accuracy.

To alleviate the issues found in the baseline models focus was directed towards reducing computational costs, minimising output generation time, increasing model generalisability, removing self-linking tendencies and maintaining high accuracy with marginal trade offs. The resulting model operates seamlessly on consumer hardware, offering a practical and efficient alternative for managing large-scale MPCs.

## III. METHODOLOGY

The proposed model incorporates various means to optimise result generation. The model is divided into distinct components as in Fig, 2, serving as a concise visual guide outlining these components. Subsequent subsections elaborate on the model's architecture and functionality.

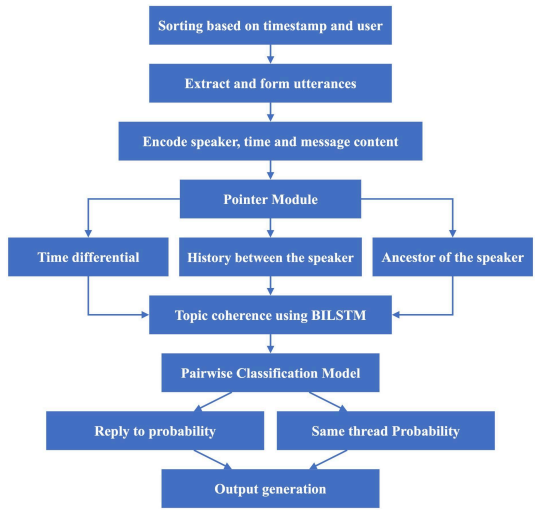

Fig. 1. Pointer Networks

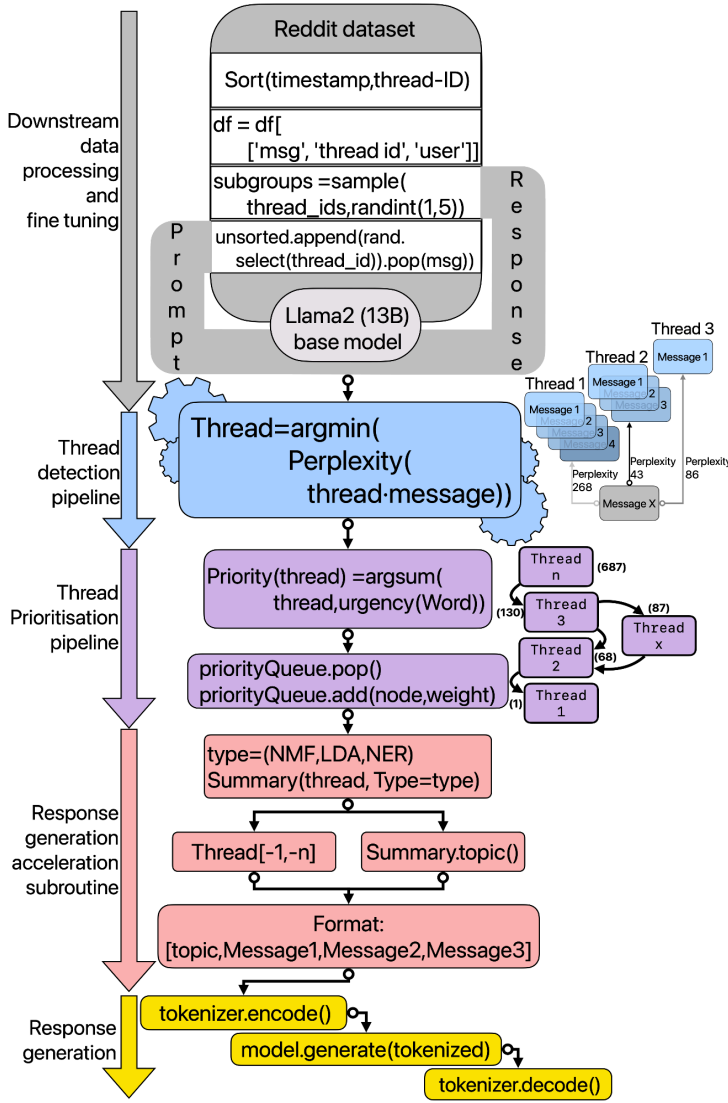

Fig .2. Proposed Llama Model

### Algorithm 1 Forming dataset:

1: **Input:** data from api endpoint
2: Set the seed for reproducibility
3: Preprocess the text data and create a dictionary of messages grouped by thread-id
4: Get unique thread_ids
5: **while** thread_ids:
6:  take random sample of 1-5 ungrouped thread_ids and add group to groups threads
7:  remove selected thread ids from thread_ids
8: **for** subset in subsets:
9:  **while** subset:
10:   select random thread from subset and pop first message
11:   append message to unsorted list
12:   **if** all messages from thread_id in unsorted_list:
13:    remove thread_id from random_subsample
14:   append message to random_subsample
15:  append random_subsample to result_subset
16: wholeset = []
17: **for** i in range(min(len(unsorted),len(sorted))):
18:  formatted_unsorted_texts = "".join(unsorted[i])
19:  formatted_sorted_texts = "".join([f"{string}" for string in sorted[i]])
20: entry=f"<s>[INST]{formatted_unsorted_texts}[/INST]{sformatted_sorted_textst}</s>"
21: wholeset.append([entry])
22: **Output:** Ordered dataset

### A. Downstream Data Processing and fine tuning:

The reddit dataset[13] used, is ideal due to the structural nature of reddit comments and metadata available. To address the need for a more naturalistic simulation of conversational environments, a novel approach was undertaken. Originally, the dataset comprised of isolated threads, distinguished by a unique thread identifier (thread id), resulting in a training scenario where the model predominantly encountered single-thread instances. Since the data was arranged in time wise order and each thread had a distinctly different time frame, which would merely arrange the threads sequentially. This was not a viable solution to directly implement cases of parallel conversation. To overcome this limitation, Algorithm 1 was used.

### B. Thread detection pipeline:

Perplexity is a measure used in natural language processing and information theory to evaluate the performance of language models [14]. It provides a quantitative assessment of how well a probability distribution or a language model predicts a given dataset.

In the context of language models, perplexity measures how surprised or uncertain a model is when predicting the next word in a sequence of words. It is calculated as the inverse probability of the test set, normalised by the number of words. Mathematically, perplexity (PPL) is defined as:

$$PPL(W) = \sqrt[N]{\frac{1}{P(w_1, w_2, w_3 \ldots w_n)}}$$

Where N is the number of words in the test set, and P(w1,w2,…,wN) is the probability assigned by the model to the entire sequence of words in the test set.

A lower perplexity indicates that the model has a higher probability of predicting the requested output. However it also signifies the fact that a response that has a higher probability/lower perplexity of being predicted may be a continuation of said prompt. If a message is a part of the thread, it would be more likely for the model to predict it as the response for the prompt (thread here). And as such it can be used as a metric to calculate if the message is a likely response of a thread. This is a much faster evaluation method as calculating perplexity does not require the model to generate the output and as such have smaller computational requirements.

*C. Thread Prioritisation Pipeline:*

A crucial factor in critical communication in a resource constrained hardware environment is the allocation of resources for computing at the right time. This means using a different prioritisation method different from a simple FIFO Queue. To implement this, a dictionary is created with each word in the dictionary being assigned a weight corresponding to the order of their importance. The total weight of the thread calculated using its urgency based keywords, and the time elapsed since the request has been sent, is used in the function to calculate the node's priority in the queue and then based on the order they are evaluated and the queue is constantly updated in real time [15].

*D. Prompt Optimization Subroutine:*

Due to the quadratic time complexity of transformer-based models, it is imperative to optimise the prompt length. Using methods like Non-negative matrix factorisation(NMF) / Latent Dirichlet Allocation(LDA) / Named Entity Recognition(NER), We can summarise the contents of the document, reducing linguistic padding induced grammatic requirements, allowing for faster evaluation. The prompt will consist of the topic and the last-n messages of the thread which will be encoded and used as input for the LLM. Subsections B, C and D are implemented together in Algorithm 2, as the output of each subsection is responsible for the next.

*E. Model Selection and Response Generation:*

The implementation leveraged the pre-trained Llama model, which has undergone extensive training on a large dataset of publicly available information. The advantage of this approach lies in Llama's ability to retain and build upon the inferences made during its pre-training phase. The Llama 7b model is specifically chosen due to the inherent nature of conversational threads, which tend to be longer, necessitating a model with a larger context length. However the exact flavour of the LLM may be changed when a better open sourced LLM arises later.

**Algorithm 2** Identify threads and generate response:

1: **Input:** individual messages
2: **Initialise:** Threads=[]
3: **for** message in messages:
4:   **for** thread in threads:
5:     perplexity_score..append(perplexity(
            "\n".join(thread+[message])))
6:   **if** min(perplexity_score)>threshold:
7:     threads.append([message])
8:   **else:**
9:     thread_idx =
          perplexity_score.index(min(perplexity_score))
10:    threads[thread_idx] = threads[thread_idx]+[message]
11:   **if** len(threads[thread_idx])>max_len:
12:     messages=threads[thread_idx]
12a:    vect = CountVectorizer()
12b:    vect = TfidfVectorizer()
13:     X = vectorizer.fit_transform(messages)
14a:    model=LatentDirichletAllocation().fit(X)
14b:    model = NMF().fit(X)
15:     feature_names = vect.get_feature_names_out()
16:     topics = " ".join([feature_names[i] for i in
             model.components_[0].argsort()[:-5:-1]])
17:     threads[thread_idx] = [topics]+messages
18: **for** thread in threads:
19:   thread_inspect="\n".join(thread)
20:   weight=0
21:   **for** word in thread_inspect:
22:     weight=weight + weightDict[word]
23:   **if** priorityQueue.exists(thread):
24:     thread.weight=weight
25: thread=PriorityQueue.pophead()
26: prompt_thread=thread[0]+thread[-1,-max_len]
27: tokenized = tokenizer.encode(prompt_thread)
28: MultiConvModel.generate(tokenized)

Given the quadratic time complexity of transformer-based models, a careful consideration of model parameters was undertaken. While a larger number of parameters could potentially enhance the model's performance, it could also lead to a significant slowdown in inference time. The decision to opt for the Llama2 7b model strikes a balance between sufficient model complexity and manageable inference time.

Moreover, the design of the model was oriented towards utilising the instruct model instead of the chat model. This strategic choice was made to avoid the fixed prompt template inherent in the chat model. By doing so, the model can be enhanced and adapted without the need to construct a transformer-based model from the ground up, affording flexibility for future improvements leveraging open source baseline models.

## IV. RESULTS AND DISCUSSION

Throughout the training and testing process, computations were performed on either a google colab t4 runtime or a computing system equipped with an Intel(R) Xeon(R) E-2236 CPU @ 3.40GHz and an NVIDIA Quadro RTX 4000 GPU. Data retrieval and storage operations were facilitated by a Samsung Electronics NVMe SSD Controller SM981/PM981 on a system running UBUNTU 20.04.3 LTS. These hardware components collectively contributed to the timely execution of our computational tasks, ensuring the efficiency and reliability of the results presented in the subsequent sections.

*A. Baseline Models:*

The pointer networks model was tested using the training dataset having a statistic of one utterance having only 1.03 parent utterances on average as per the Ubuntu data excluding self links (the drawback of this model) and another including self links. The model performed significantly well with the first dataset with an accuracy rate of 94%, performing well on the factors of link prediction due to the fact that in pointer networks, only the significant information of timestamp and message information rather than collecting all the metadata, thereby handling the cases of absence of messages and topic coherence way better. The dataset with self links comprising 40% of the conservation showed a significant drop to 75% due to the self link being predicted wrong and utterances being pointed to wrong parts of the conversation and destroying the two previously positive clusters.

This bottleneck present in the pointer networks model is effectively tackled by the proposed Llama model by reducing the pointer network to operate within the transformers which handles long range dependencies sequentially without necessitating self links by utilising self-attention. Due to the characteristics of transformer models, the graphs are internalised by multihead attention, which reduces the extra computational overhead required by the pointer network.

Fig. 3. Pointer Networks Result

Fig. 4. Results of Llama Model

*B. Proposed Model*

The Llama2 Transformer based model, following the fine tuning approach, proves to be more context aware and with responses being closely related to the conversation by analysing the model's relevance completing prompt. The model takes conversation as the prompt and suggests a continuation as per the context size after identifying the scenario. The model addresses a wide variety of conversation types including, conversation of two people on same topic, conversation of more than two people on same topic, conversation of two people on different topics, and conversation of more than two people on different topics.

Unlike other models it is much more efficient and consistent in generating natural responses in a normal conversation like scenario. The Llama model provides a response with contextual relevance at least in ~90% where even contemporary models fail to generate outputs. Response time is up to 10X faster than other contemporary models with comparable accuracy in response generation in ~3 minutes as compared to ~30 minutes. It proves its usability as a conversational agent to suggest probable responses to users.

## V. CONCLUSION

The proposed model eliminates the use of graphs in the analysis of MPC. The model has been optimised to work on consumer grade hardware by implementing prompt refactoring techniques thereby lowering the processing requirements. The model achieves up to 10x speed improvement, while generating more coherent results compared to existing models, showcasing its potential for real-world scenarios.

While the model demonstrates versatility, it is important to note that the results are generalised. Further fine tuning and exploration is warranted to efficiently implement the model for domain-specific applications or nuanced conversations. This strategic enhancement could significantly elevate the model's performance in scenarios where contextual understanding plays a pivotal role.

Through continued research and development, we anticipate not only addressing specific domain intricacies but also advancing the overall efficiency and adaptability of our model, widening the application domain for the models.

# REFERENCES

[1] Li, Jiaqi, et al. "Molweni: A challenge multiparty dialogues-based machine reading comprehension dataset with discourse structure." arXiv preprint arXiv:2004.05080 (2020).

[2] Ma, Xinbei, Zhuosheng Zhang, and Hai Zhao. "Structural characterization for dialogue disentanglement." arXiv preprint arXiv:2110.08018 (2021).

[3] Meng, Zhao, Lili Mou, and Zhi Jin. "Towards neural speaker modeling in multi-party conversation: The task, dataset, and models." Proceedings of the AAAI Conference on Artificial Intelligence. Vol. 32. No. 1. 2018.

[4] Plank, Barbara, Anders Søgaard, and Yoav Goldberg. "Multilingual part-of-speech tagging with bidirectional long short-term memory models and auxiliary loss." arXiv preprint arXiv:1604.05529 (2016).

[5] Reimers, Nils, and Iryna Gurevych. "Sentence-bert: Sentence embeddings using siamese bert-networks." arXiv preprint arXiv:1908.10084 (2019).

[6] Lai, Tuan Manh, et al. "A simple but effective bert model for dialog state tracking on resource-limited systems." ICASSP 2020-2020 IEEE International Conference on Acoustics, Speech and Signal Processing (ICASSP). IEEE, 2020.

[8] Chao, Guan-Lin, and Ian Lane. "BERT-DST: Scalable end-to-end dialogue state tracking with bidirectional encoder representations from transformer." arXiv preprint arXiv:1907.03040 (2019).

[8] Li, Tianda, et al. "Dialbert: A hierarchical pre-trained model for conversation disentanglement." arXiv preprint arXiv:2004.03760 (2020).

[9] Yu, Tao, and Shafiq Joty. "Online conversation disentanglement with pointer networks." arXiv preprint arXiv:2010.11080 (2020).

[10] Gu, Jia-Chen, et al. "Hetermpc: A heterogeneous graph neural network for response generation in multi-party conversations." arXiv preprint arXiv:2203.08500 (2022).

[11] Liu, Hui, et al. "End-to-End Transition-Based Online Dialogue Disentanglement." IJCAI. Vol. 20. 2020.

[12] Touvron, Hugo, et al. "Llama 2: Open foundation and fine-tuned chat models." arXiv preprint arXiv:2307.09288 (2023).

[13] Tan, Ming, et al. "Context-aware conversation thread detection in multi-party chat." Proceedings of the 2019 Conference on Empirical Methods in Natural Language Processing and the 9th International Joint Conference on Natural Language Processing (EMNLP-IJCNLP). 2019.

[14] Meister, Clara, and Ryan Cotterell. "Language model evaluation beyond perplexity." arXiv preprint arXiv:2106.00085 (2021).

[15] Islam, Azharul, and KyungHi Chang. "Real-time AI-based informational decision-making support system utilizing dynamic text sources." Applied Sciences 11.13 (2021): 6237.